\documentclass[a4paper,fleqn]{cas-dc}

\usepackage[round, sort]{natbib}
\usepackage{microtype}
\usepackage{placeins}
\usepackage{amsmath}
\usepackage{tabularx}
\usepackage{array}

\begin{document}
\let\WriteBookmarks\relax
\def\floatpagepagefraction{1}
\def\textpagefraction{.001}
\shorttitle{One-Time Training for All Grains}
\shortauthors{Su et~al.}

\title [mode = title]{One-Time Training for All Grains: Open-Set Grain Recognition and Quantitative Analysis
}                      
\author[2,3]{Qihe Su}
\fnmark[1]
\credit{Conceptualization, Methodology, Software, Data curation, Formal analysis, Visualization, Writing -- original draft}

\author[2,4]{Mengyu Sun}
\fnmark[1]
\credit{Investigation, Data curation, Validation, Writing -- review and editing}

\author[2,3]{Yuxi Ke}
\credit{Data curation, Investigation, Conceptualization}

\author[2,3]{Zhuoyan Jiang}
\credit{Data curation, Investigation}

\author[1,3]{Wanneng Yang}
\credit{Resources, Project administration}

\author[1,2,3]{Chenglong Huang}
\cormark[1]
\ead{hcl@mail.hzau.edu.cn}
\credit{Supervision, Project administration, Resources, Funding acquisition, Writing -- review and editing}

\author[4]{Ziyuan Yang}
\cormark[1]
\ead{cziyuanyang@gmail.com}
\credit{Methodology, Validation, Resources, Writing -- review and editing}


\affiliation[1]{
    organization={National Key Laboratory of Crop Genetic Improvement, National Center of Plant Gene Research, Hubei Hongshan Laboratory, Huazhong Agricultural University},
    city={Wuhan},
    postcode={430070},
    country={China}
}

\affiliation[2]{
    organization={College of Engineering, Huazhong Agricultural University},
    city={Wuhan},
    postcode={430070},
    country={China}
}

\affiliation[3]{
    organization={Engineering Research Center of Intelligent Technology for Agriculture, Ministry of Education}
}

\affiliation[4]{
    organization={School of Cyber Science and Engineering, Sichuan University},
    city={Chengdu},
    postcode={610207},
    country={China}
}
\cortext[cor1]{Corresponding author}
\fntext[fn1]{These authors contributed equally to this work.}

\begin{abstract}
Advances in crop breeding have introduced an increasing number of grain varieties, creating a growing demand for efficient variety recognition and quantitative analysis. However, existing methods are typically trained on a fixed variety set, and incorporating newly introduced varieties requires additional data collection and model retraining. To address this limitation, we propose \textbf{GROW}, a framework for \textbf{G}rain \textbf{R}ecognition and quantitative analysis in \textbf{O}pen sets \textbf{W}ithout retraining. GROW first performs class-agnostic grain localization, converting mixed-grain images into individual instances for variety-wise counting and phenotypic measurement. It then combines visual embeddings and morphological
descriptors into fused grain descriptors stored in an extensible \textbf{GrainBank}. Query grains are recognized through rank--similarity weighted top-$k$ retrieval, and newly introduced varieties are incorporated by appending their descriptors without updating the deployed models.
Extensive experiments under progressive variety expansion, varying grain densities, and background domain shifts demonstrate the scalability, robustness, and adaptability of GROW. Compared with joint retraining, GROW reduced the average category-registration time from 4153~s to only 39~s while maintaining competitive recognition performance. These results demonstrate that GROW provides an efficient and maintainable solution for extensible grain recognition, counting, and phenotypic analysis without repeated model retraining.
\end{abstract}

\begin{highlights}

\item GROW shifts novel-variety expansion from model retraining to rapid GrainBank updating.

\item Class-agnostic localization decouples grain detection from variety inference in mixed-grain images.

\item GROW maintains balanced base- and novel-variety recognition while supporting background recovery through GrainBank updates.

\end{highlights}


\maketitle

\section{Introduction}
Advances in grain breeding have introduced an increasing diversity of varieties, placing greater demands on automated grain recognition, counting, and phenotypic measurement in grain sorting and cultivar evaluation \citep{velesacaComputerVisionBased2021,Vithu,sun2023novel}.

Conventional grain analysis relies heavily on manual inspection and physical measurement, which are labor-intensive, time-consuming, and sensitive to operator experience, particularly when large numbers of breeding materials must be examined. 
Machine vision provides a non-contact, repeatable, and high-throughput alternative for automated grain recognition, counting, and phenotypic measurement \citep{Liu,Mahajan,ghazalComputerVision2024}.
Recent advances in deep learning have further improved grain counting, phenotypic measurement, and variety recognition under increasingly complex imaging conditions \citep{kamilarisDeepLearningAgriculture2018,laabassiWheatVarietiesIdentification2021,wangGrainNetEfficientDetection2025,oufLeguminousSeeds2023}.
Despite their robustness to variations in background, scale, and grain density, most existing methods are developed for a fixed variety set, assuming that all varieties encountered during deployment have already been observed during training\citep{velesacaComputerVisionBased2021, rajalakshmiRiceSeedNetRiceSeed2024}. 
Their recognition capability is therefore restricted to predefined output categories.
When a novel variety is introduced, its incorporation generally requires collecting and annotating a large number of grain images and retraining the model on existing-variety data \citep{wangGrainNetEfficientDetection2025}.
Repeating this process consumes substantial data and computational resources\citep{xu2025low}, prolongs the deployment cycle, and becomes increasingly difficult to maintain as the variety set expands \citep{maWheatSeedDetection2024a,zouRiceGrainDetection2023}.
Class-incremental learning provides a possible alternative by updating a trained model with newly introduced varieties. Existing approaches commonly employ parameter regularization, knowledge distillation, rehearsal, or stored exemplars to preserve previously acquired knowledge \citep{kirkpatrickOvercomingCatastrophicForgetting2017,liLearningWithoutForgetting2018,rebuffiICaRL2017,parisiContinualLifelongLearning2019, Zhou_2024}. 

Nevertheless, these methods still modify model parameters whenever novel varieties are introduced. Each update acquire knowledge of the novel varieties while preserving recognition of the existing ones, making the process susceptible to catastrophic forgetting, representation drift, and data imbalance.
Consequently, the practical requirement is not merely higher performance on a fixed variety set, but the ability to extend a deployed grain analysis system to novel varieties with minimal data and computation while preserving its existing capabilities.

\begin{figure}
    \centering
    \includegraphics[width=1\linewidth]{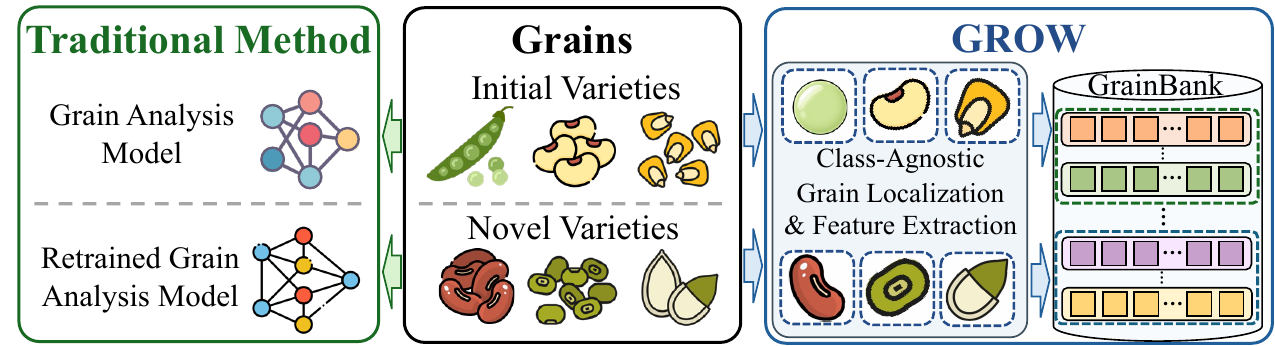}
    \caption{Comparison between conventional methods and GROW.}
    \label{fig1}
\end{figure}

Based on this perspective, we propose \textbf{GROW}, a framework for \textbf{G}rain \textbf{R}ecognition and quantitative analysis in \textbf{O}pen sets \textbf{W}ithout retraining. 
As illustrated in Fig.~\ref{fig1}, conventional methods perform grain recognition and quantitative analysis within a predefined variety set and require model retraining when novel varieties are introduced. In contrast, GROW first transforms mixed-grain images into individual grain instances for counting and phenotypic measurement, and then recognizes their varieties through feature matching with the \emph{GrainBank}.

Specifically, GROW first treats all varieties as a single grain category to simplify grain localization, improve counting robustness, and remain applicable to newly introduced varieties. It then constructs an extensible GrainBank that stores grain features extracted from each variety.
During inference, each query grain is recognized by matching its feature against those stored in the GrainBank. Novel varieties can be incorporated by adding their features to the GrainBank using only a small set of grain images, without retraining the deployed model.
We evaluate GROW under progressive variety expansion, varying grain densities, and background shifts. The results demonstrate its scalability and robustness for grain recognition and quantitative analysis under challenging open-set conditions.

Our main contributions are summarized as follows:
\begin{itemize}
\item We propose \textbf{GROW}, an open-set grain analysis paradigm that replaces fixed-category variety classification with class-agnostic individual-grain localization and GrainBank-based feature matching, supporting reliable counting and phenotypic measurement while enabling novel varieties to be incorporated without model retraining.
\item We introduce class-agnostic grain localization, where all varieties are treated as a single grain category. Compared with multi-category detection, it improves grain localization and counting robustness across different imaging conditions while remaining applicable to newly introduced varieties.
\item We construct an extensible GrainBank that stores grain features from different varieties. At inference, query grains are recognized through feature matching with the GrainBank. Novel varieties are incorporated by adding features from a small set of grain images without retraining the deployed model.

\end{itemize}


\section{Materials and methods}
\subsection{Problem Statement}
\label{sec:problem_statement}

\begin{figure*}
    \centering
    \includegraphics[width=\linewidth]{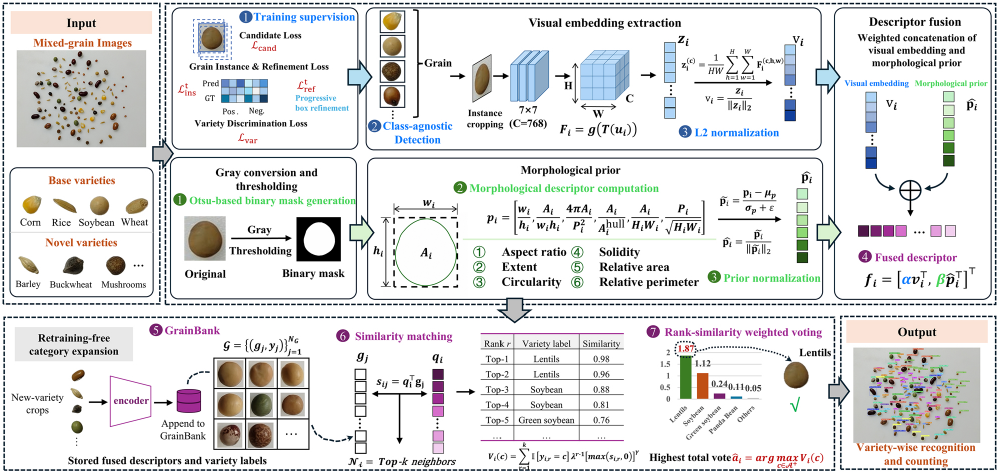}
    \caption{Overview of the proposed GROW framework.}
    \label{fig2}
\end{figure*}

Grain recognition in mixed images involves locating individual grain instances
and determining their corresponding varieties. Let $\mathbf{x}$ denote a
mixed-grain image, and let $\mathcal{A}=\{a_1,\ldots,a_M\}$ denote the set of
$M$ grain varieties available during training. Conventional methods learn a
recognition model $F_{\boldsymbol{\theta}}$ from these varieties, where
$\boldsymbol{\theta}$ denotes the model parameters. Given $\mathbf{x}$, the
model directly predicts the locations and varieties of the detected grains:
\begin{equation}
    F_{\boldsymbol{\theta}}(\mathbf{x})
    =
    \left\{
    \left(\hat{r}_i,\hat{c}_i\right)
    \right\}_{i=1}^{\hat{N}},
    \qquad
    \hat{c}_i \in \mathcal{A},
\end{equation}
where $\widehat{\mathbf r}_i$ and $\widehat c_i$ denote the predicted bounding-box region and variety of the $i$-th grain, respectively, and $\hat{N}$ denotes the number of detected grains. 
Since the output categories of $F_{\boldsymbol{\theta}}$ are defined by $\mathcal{A}$, grains belonging to varieties outside this set cannot be assigned to their corresponding varieties. 
When an additional variety set $\mathcal B=\{b_1,\ldots,b_K\}$ of $K$ varieties is introduced, the expanded variety set is denoted by $\mathcal{A}^{+}=\mathcal{A}\cup\mathcal{B}$. Conventional methods therefore require additional data collection and model retraining to recognize varieties in $\mathcal{A}^{+}$.

To address this limitation, we reformulate grain recognition from single-step direct prediction into a two-step process of class-agnostic grain localization and feature matching against an extensible GrainBank.
The GrainBank is initialized with descriptors extracted from grains of the varieties in $\mathcal A$. When the varieties in $\mathcal{B}$ are introduced, their features can be added to the GrainBank using only a small number of images. 
Given the fused descriptor $\mathbf q_i$ of the $i$-th localized query
grain and the GrainBank $\mathcal G$, its predicted variety is obtained by:
\begin{equation}
\widehat c_i
=
\arg\max_{c\in\mathcal A^{+}}
S_i(c;\mathcal G),
\end{equation}
where $S_i(c;\mathcal G)$ denotes the accumulated matching score of
candidate variety $c$ under the globally ranked GrainBank entries, as
defined in Section~\ref{sec:grainbank_inference}. In this way, newly introduced varieties can be recognized by updating the GrainBank rather than retraining the deployed models.

\subsection{Overview}
An overview of GROW is illustrated in Fig.~\ref{fig2}. GROW consists of three sequential stages: (1) class-agnostic grain localization, (2) variety-discriminative visual feature extraction and morphological
trait measurement, and (3) GrainBank-based variety inference. For each localized grain, GROW extracts a variety-discriminative visual embedding and a morphological descriptor, which are combined by weighted concatenation to form a fused grain descriptor. The fused descriptors of the initial varieties are then stored in the GrainBank. During inference, each query descriptor is matched against the stored descriptors, and its variety is determined through rank--similarity weighted top-$k$ voting. When a new variety is introduced, it can be incorporated into the GrainBank using only a small number of images, without retraining the localization model or grain encoder.

\subsection{Class-Agnostic Grain Localization}
We employ a class-agnostic Grain Localization Network $\mathcal{D}_{\boldsymbol{\phi}}$, where $\boldsymbol{\phi}$ denotes its parameters. During training, grains from all varieties are treated as a common foreground category, enabling the localization model to remain applicable when the variety
set is expanded. Given a mixed-grain image $\mathbf{x}$, the model localizes individual grain instances, thereby reformulating multi-grain recognition as single-grain analysis and improving robustness to variations in grain density, overlap, and spatial arrangement.

The training of $\mathcal{D}_{\boldsymbol{\phi}}$ is supervised by the Grain Candidate Loss $\mathcal{L}_{\mathrm{cand}}$, which supervises the generation of initial candidate grain regions, together with the Grain Instance Loss $\mathcal{L}_{\mathrm{ins}}^{t}$ and Grain Refinement Loss $\mathcal{L}_{\mathrm{ref}}^{t}$ at each refinement stage. The overall objective is formulated as:
\begin{equation}
\mathcal{L}_{\mathrm{GL}} = \mathcal{L}_{\mathrm{cand}} +
\sum_{t=1}^{T}
\left(
\mathcal{L}_{\mathrm{ins}}^{t} +
\lambda_{\mathrm{ref}}
\mathcal{L}_{\mathrm{ref}}^{t}
\right),
\label{eq:grain_localization_objective}
\end{equation}
where the losses at the $t$-th refinement stage are defined as:
\begin{align}
\mathcal{L}_{\mathrm{ins}}^{t} &=
-\frac{1}{N_t}
\sum_{k=1}^{N_t}
\left[
z_k^{t}\log q_k^{t} +
\left(1-z_k^{t}\right)
\log\left(1-q_k^{t}\right)
\right],\\
\mathcal{L}_{\mathrm{ref}}^{t} &=
\frac{1}{N_t^{+}}
\sum_{k=1}^{N_t}
z_k^{t}
\left\|
\widehat{\mathbf{r}}_k^{t} - \mathbf{r}_k
\right\|_1 .
\label{eq:grain_localization_terms}
\end{align}

Here, $T$ denotes the number of refinement stages. At the $t$-th stage, $N_t$ is the number of processed regions, $q_k^{t}$ is the predicted probability that the $k$-th region contains a grain, and $z_k^{t}\in\{0,1\}$ is determined by the overlap between the region and its matched annotation.
$N_t^{+}=\sum_{k=1}^{N_t}z_k^{t}$ is the number of positive regions, while $\widehat{\mathbf{r}}_k^{t}$ and $\mathbf{r}_k$ denote the predicted and annotated region parameters, respectively.
$\lambda_{\mathrm{ref}}$ controls the contribution of the Grain Refinement Loss. The matching threshold is increased across successive stages, allowing the localized regions to be progressively improved.

Given a mixed-grain image $\mathbf{x}$, the trained localization model produces a set of individual grain regions
$\mathcal{D}_{\boldsymbol{\phi}}(\mathbf{x}) =
\{(\widehat{\mathbf{r}}_i,s_i)\}_{i=1}^{\widehat{N}}$,
where $\widehat{\mathbf{r}}_i$ and $s_i$ denote the localized region and confidence score of the $i$-th grain, respectively, and $\widehat{N}$ is the number of retained regions. Each retained region is cropped from $\mathbf x$, resized to a fixed spatial resolution, and normalized to obtain the corresponding single-grain image $\mathbf u_i$.

\subsection{Variety-Discriminative Grain Feature Extraction}
Using the localized grain images $\mathcal{U}_{a}=\{\mathbf{u}_{a,n}\}_{n=1}^{N_a}$ from all initial varieties $a\in\mathcal{A}$, where $\mathbf{u}_{a,n}$ denotes the $n$-th grain image from variety $a$, we jointly train a grain encoder $\mathcal{E}_{\boldsymbol{\psi}}$ and a variety discrimination head $\mathcal{H}_{\boldsymbol{\omega}}$, where $\boldsymbol{\psi}$ and $\boldsymbol{\omega}$ denote their parameters, respectively.
\begin{equation}
\mathcal{L}_{\mathrm{var}} =
\left(
\sum_{a\in\mathcal{A}}N_a
\right)^{-1}
\sum_{a\in\mathcal{A}}
\sum_{n=1}^{N_a}
\ell_{\mathrm{ce}}
\left(
\mathcal{H}_{\boldsymbol{\omega}}
\left(
\mathcal{E}_{\boldsymbol{\psi}}(\mathbf{u}_{a,n})
\right), a\right),
\label{eq:variety_loss}
\end{equation}
where $N_a$ denotes the number of localized grain images from variety $a$, and $\ell_{\mathrm{ce}}$ is the cross-entropy loss. 

Through the variety discrimination head $\mathcal H_{\boldsymbol\omega}$, this objective supervises $\mathcal E_{\boldsymbol\psi}$ to extract visual features that distinguish the initial varieties. 
After training, $\mathcal{E}_{\boldsymbol{\psi}}$ is used to extract the visual embedding:
\begin{equation}
\mathbf{v}_{a,n}
=
\mathcal{E}_{\boldsymbol{\psi}}(\mathbf{u}_{a,n}),
\qquad
\mathbf{v}_{a,n}\in\mathbb{R}^{d_v}.
\label{eq:visual_feature}
\end{equation}

In addition to the visual embedding, six morphological traits are derived from each grain image. Specifically, $\mathbf{u}_{a,n}$ is converted to grayscale, and its foreground region is segmented using Otsu thresholding \citep{otsuThresholdSelectionMethod1979}. Morphological opening and closing are applied to remove isolated noise and improve contour continuity, after which the largest connected contour is selected as the grain region. 
Based on this contour, the morphological traits are calculated as:
\begin{equation}
\begin{aligned}
\mathbf{p}_{a,n}
={}&
\Bigl[
\frac{w_{a,n}}{h_{a,n}},
\frac{A_{a,n}}{w_{a,n}h_{a,n}},
\frac{4\pi A_{a,n}}{P_{a,n}^{2}},
\\[-0.2em]
&\quad
\frac{A_{a,n}}{A_{a,n}^{\mathrm{hull}}},
\frac{A_{a,n}}{H_{a,n}W_{a,n}},
\frac{P_{a,n}}{\sqrt{H_{a,n}W_{a,n}}}
\Bigr]^{\top}
\in\mathbb{R}^{6},
\end{aligned}
\label{eq:morphological_traits}
\end{equation}
where the six components correspond to aspect ratio, extent, circularity, solidity, relative area, and relative perimeter, respectively. Here, $w_{a,n}$ and $h_{a,n}$ denote the width and height of the bounding rectangle of the grain contour; $A_{a,n}$ and $P_{a,n}$ denote its area and perimeter;
$A_{a,n}^{\mathrm{hull}}$ denotes the convex-hull area; and $H_{a,n}$ and $W_{a,n}$ denote the height and width of $\mathbf{u}_{a,n}$. These traits quantify the shape and contour characteristics of individual grains and are also reported as quantitative phenotypic data.

Before being combined with the visual embedding, the morphological traits are standardized as:
\begin{equation}
\widetilde{\mathbf{p}}_{a,n}
=
\frac{
\mathbf{p}_{a,n}-\boldsymbol{\mu}_{p}
}{
\boldsymbol{\sigma}_{p}+\varepsilon
},
\label{eq:morphological_standardization}
\end{equation}
where $\boldsymbol{\mu}_{p}$ and $\boldsymbol{\sigma}_{p}$ denote the element-wise mean and standard deviation estimated from the initial grain set, respectively, and $\varepsilon$ is a small constant for
numerical stability. These statistics remain fixed during subsequent GrainBank updates and query inference.

The visual embedding and standardized morphological descriptor are then combined by weighted concatenation to form the fused grain descriptor:
\begin{equation}
\mathbf{f}_{a,n}
=
\left[
\alpha\mathbf{v}_{a,n}^{\top},
\,
\beta\widetilde{\mathbf{p}}_{a,n}^{\top}
\right]^{\top},
\qquad
\mathbf{f}_{a,n}\in\mathbb{R}^{d_v+6},
\label{eq:grain_feature}
\end{equation}

where $\alpha\geq0$ and $\beta\geq0$ are scaling coefficients that balance the visual embedding and standardized morphological descriptor, respectively. The resulting fused grain descriptor is stored in the GrainBank for subsequent variety inference, while the six morphological traits are preserved for quantitative phenotypic analysis.

\subsection{GrainBank-based Variety Inference}
\label{sec:grainbank_inference}
The GrainBank stores fused grain descriptors together with their variety labels. For each initial variety $a\in\mathcal{A}$, the descriptors $\{\mathbf{f}_{a,n}\}_{n=1}^{N_a}$ are inserted into the
GrainBank. When a new variety $b$ is introduced, its grain images are processed using the fixed grain encoder and the same morphological descriptor construction procedure, and the resulting descriptors $\{\mathbf{f}_{b,m}\}_{m=1}^{N_b}$ are appended with label $b$, where $N_b$ is the number of reference-grain descriptors available for the newly introduced variety $b$. Thus, variety expansion requires only a GrainBank update, without modifying the trained localization model or grain encoder.

The current GrainBank is denoted by $\mathcal{G}=\{(\mathbf{g}_j,y_j)\}_{j=1}^{N_G}$, where $\mathbf{g}_j$ and $y_j$ are the fused descriptor and variety label of the $j$-th stored grain, respectively, and $N_G$ is the total number of stored descriptors.

Given a query mixed-grain image, the localized grain images are processed using the same feature extraction procedure to obtain query fused descriptors $\{\mathbf{q}_i\}_{i=1}^{\widehat{N}}$. Each query
feature is compared with every descriptor in the GrainBank to obtain the similarity score ${s_{ij}}=\mathbf{q}_i^{\top}\mathbf{g}_j$. 

For each query grain, all GrainBank entries are ranked in descending order of similarity, and the top-$k$ entries are retained as its nearest neighbors. Let $s_{i,r}$ and $y_{i,r}$ denote the similarity score and variety label of the $r$-th ranked neighbor of the $i$-th query grain, respectively.

Each retrieved neighbor contributes a voting weight determined jointly by its retrieval rank and similarity score. The contributions of neighbors with the same variety label are accumulated, and the variety with the highest total voting score is assigned to the query grain:
\begin{equation}
\widehat{c}_i
=
\arg\max_{c\in\mathcal{A}^{+}}
\sum_{r=1}^{k}
\mathbb{I}\!\left[y_{i,r}=c\right]
\lambda^{r-1}
\left[
\max\left(s_{i,r},0\right)
\right]^{\gamma},
\label{eq:grainbank_inference}
\end{equation}
where $\mathcal{A}^{+}$ denotes the current variety set in the GrainBank, including both the initial and newly introduced varieties; $\mathbb I[\cdot]$ is the indicator function; $k$ is the number of retrieved neighbors; $r$ denotes the neighbor
rank; $\lambda\in(0,1]$ controls the rank-decay weight; and $\gamma>0$ controls the contribution of neighbor similarity. The same inference procedure is applied after each GrainBank update without retraining the localization model or grain encoder.

The number of grains assigned to variety $c$ is then obtained by aggregating the instance-level predictions:
\begin{equation}
\widehat{C}_{c}
=
\sum_{i=1}^{\widehat{N}}
\mathbb{I}
\left[
\widehat{c}_{i}=c
\right],
\qquad
c\in\mathcal{A}^{+},
\label{eq:variety_counting}
\end{equation}
where $\widehat C_c$ is the predicted number of grains belonging to variety $c$. Together with the phenotypic traits extracted from each grain, these instance-level assignments provide variety-wise counting and quantitative grain
analysis.

\section{Result}

\subsection{Dataset construction and task definition}
\subsubsection{Image collection}
\begin{figure}
    \centering
    \includegraphics[width=\linewidth]{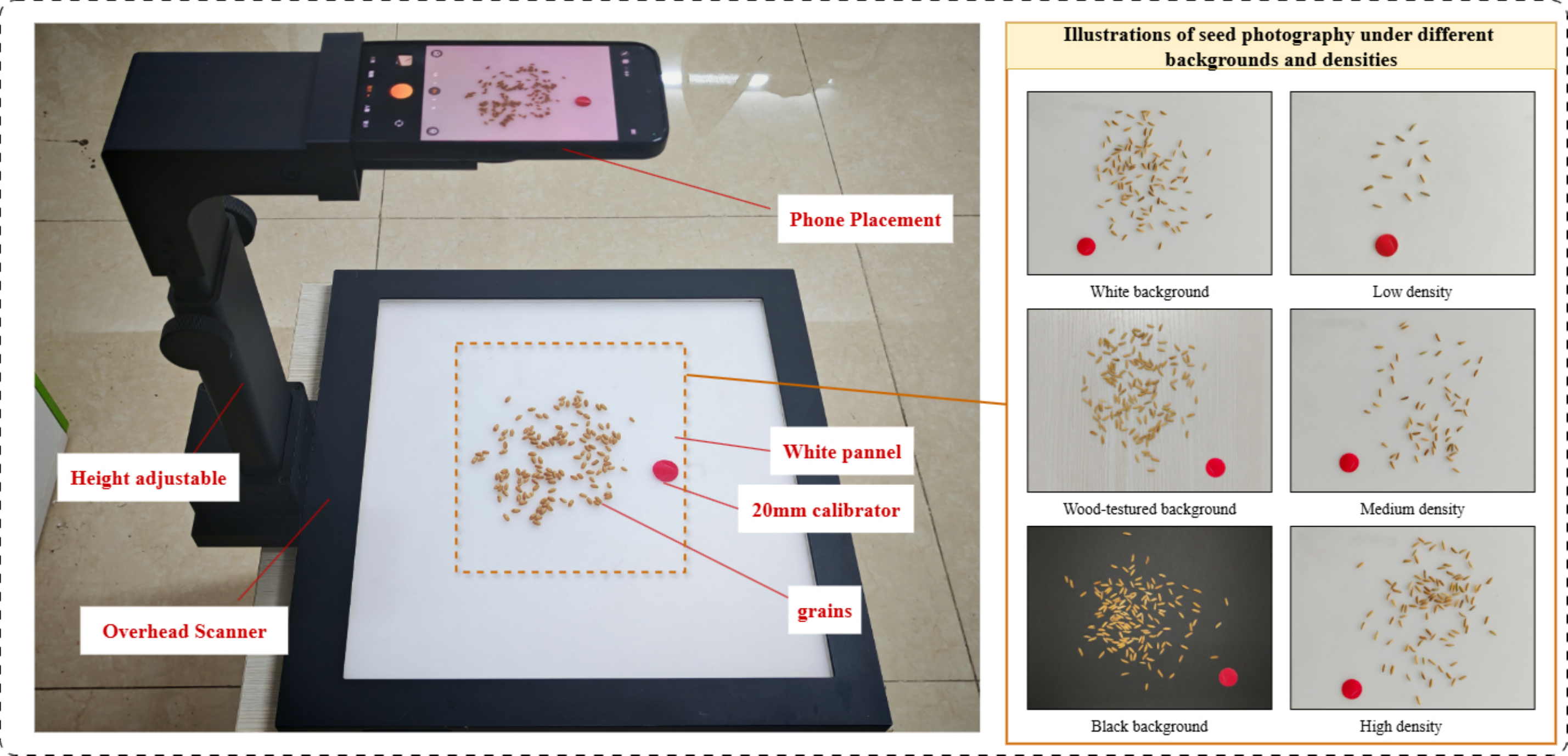}
    \caption{Grain image acquisition platform and samples with varied backgrounds and grain densities.}
    \label{fig3}
\end{figure}

Different varieties of Corn, rice, soybean, and wheat were randomly selected as experimental materials to construct the initial variety grain dataset. Grain images were collected under a backlit panel with different kernel densities. In this study, OPPO Reno8 and OPPO Find X8 smartphones were adopted for image acquisition. The dataset contains 2,800 images of initial varieties and 1,020 images of novel varieties, which were captured using the above two mobile devices, respectively. All images were captured under the environment illustrated in Fig.~\ref{fig3}. The smartphones were fixed on a high-speed document scanner. All collected images have a resolution of 3072 × 4096 and are saved in .jpg format. A red calibration circle with a diameter of 20 mm was placed within the shooting range to facilitate acquiring the real size of grains in real time.

\subsubsection{Data annotation and dataset division}

In this study, the dataset was annotated using LabelImg software. Specifically, all images were labeled via a combination of manual and semi-automatic annotation. For the training of the detection head, the dataset was randomly divided into training, validation, and test sets with a ratio of 7:2:1. The encoder was trained using 2800 images of four original grain types. Furthermore, the 1020 novel-variety grain images were used to evaluate the retraining-free variety expansion. Concretely, GrainBank was constructed using four types of grains, namely corn, rice, soybean, and wheat. New grain categories were subsequently incorporated to validate the reliability of the proposed framework.

\subsubsection{Implement detail}

The proposed framework was implemented in PyTorch on a desktop workstation equipped with an Intel Core i7-14790F CPU, an NVIDIA RTX 5070 GPU with 12 GB memory, and 32 GB RAM. The original mixed-grain images were processed at full image resolution during the detection stage, whereas each detected grain instance was cropped and then padded to a square shape with a white background before being resized to 224×224 pixels for ConvNeXt-based visual embedding extraction \citep{liuConvNet2022}. Cascade R-CNN \citep{8578742} was used as the class-agnostic grain localization network, and the subsequent grain localization was decoupled from variety inference. For each cropped grain instance, a 768-dimensional visual embedding was extracted by the ConvNeXt backbone, and a 6-dimensional morphological descriptor vector was further computed and fused with the visual embedding to form the fused grain descriptor used for GrainBank-based retrieval.The hyperparameters $k$=5, $\lambda$=0.75, $\gamma$=1.0, $\alpha$=1.0, and $\beta$=0.2 were selected using the validation set and were fixed for all subsequent experiments. All experiments were conducted under the same hardware and software environment to ensure fair comparison across different settings.

\subsection{Experimental protocol and evaluation metrics}
\label{sec:experimental_protocol}

For systematic performance assessment under retraining-free variety expansion, experiments were organized around five aspects: mixed-grain category complexity, grain density, variety-wise degradation, background domain shift, and GrainBank construction strategy. All experiments used the localization model and grain encoder were kept fixed to ensure fair comparison.

The evaluation involved four base grain varieties and thirteen newly introduced varieties. A total of 1020 annotated test images were used, including 170 mixed-grain images and 850 single-variety images. Among the single-category images, 510 images were collected under three grain-density levels on a white background, and 340 images were collected on black and wood-textured backgrounds. The density levels were defined according to the approximate number of grains per image: low density contained about 10 grains, medium density contained 10--50 grains, and high density contained more than 50 grains.

The experimental protocol was designed to assess both recognition scalability and robustness under practical variations. Mixed-category experiments were used to evaluate progressive GrainBank expansion, density experiments were used to analyze localization and retrieval stability, variety-wise degradation experiments quantified category-specific robustness, and background experiments assessed cross-domain generalization and GrainBank adaptation.

\subsection{Recognition scalability under progressive GrainBank expansion}

\begin{figure}
    \centering
    \includegraphics[width=\linewidth]{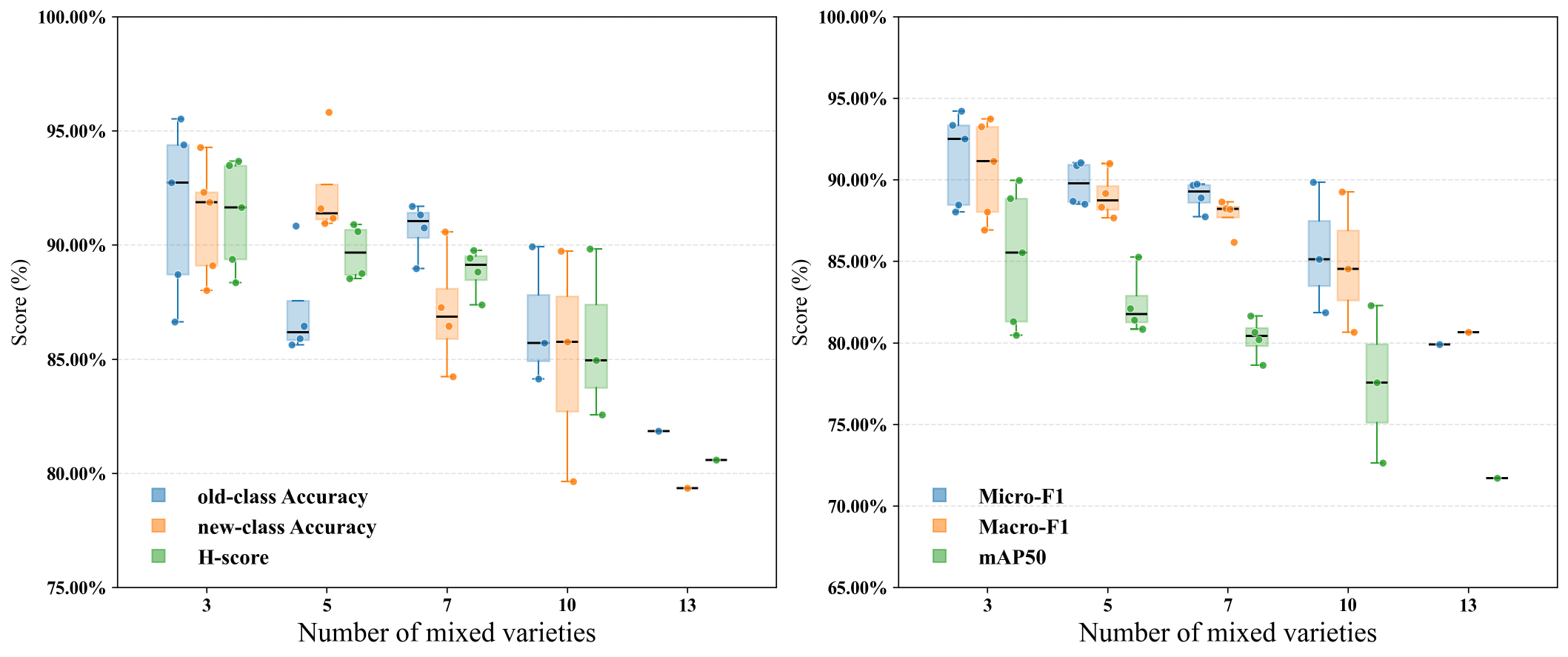}
    \caption{Overall recognition performance during progressive GrainBank expansion with increasing mixed-grain variety-set size.}
    \label{fig4}
\end{figure}

\begin{table*}[t]
\centering
\caption{Overall recognition performance under increasing mixed-variety complexity.}
\label{tab1}
\resizebox{\textwidth}{!}{
\begin{tabular}{cccccccc}
\hline
Added varieties & $n$ & Initial Acc. (\%) & Novel Acc. (\%) & H-score (\%) & Micro-F1 (\%) & Macro-F1 (\%) & mAP@50 (\%) \\
\hline
3  & 5 & $91.59 \pm 3.79$ & $91.11 \pm 2.53$ & $91.31 \pm 2.39$ & $91.30 \pm 2.85$ & $90.61 \pm 3.05$ & $85.22 \pm 4.29$ \\
5  & 4 & $87.21 \pm 2.44$ & $92.38 \pm 2.29$ & $89.69 \pm 1.22$ & $89.78 \pm 1.36$ & $89.04 \pm 1.44$ & $82.39 \pm 1.98$ \\
7  & 4 & $90.69 \pm 1.21$ & $87.13 \pm 2.62$ & $88.85 \pm 1.05$ & $89.01 \pm 0.93$ & $87.81 \pm 1.12$ & $80.29 \pm 1.25$ \\
10 & 3 & $86.59 \pm 2.99$ & $85.04 \pm 5.08$ & $85.78 \pm 3.70$ & $85.60 \pm 4.02$ & $84.82 \pm 4.31$ & $77.49 \pm 4.83$ \\
13 & 1 & 81.85 & 79.35 & 80.58 & 79.91 & 80.66 & 71.69 \\
\hline
\end{tabular}}
\end{table*}

This experiment was designed to evaluate the recognition scalability of GROW under progressive GrainBank expansion. As shown in Fig.~\ref{fig4}, the overall performance gradually decreased as more grain varieties were introduced. This trend can be attributed to the simultaneous expansion of the candidate variety space and the resulting increase in inter-variety competition, which naturally raises the difficulty of feature retrieval. To examine the sensitivity of the results to different variety combinations, multiple evaluations were conducted at the 3-, 5-, 7-, and 10-variety expansion stages using randomly selected combinations of newly introduced varieties. As reported in Table~\ref{tab1}, although the evaluated metrics exhibited varying degrees of gradual degradation, the results of the repeated evaluations remained relatively concentrated.

A comparison between the accuracies of the base and newly introduced varieties further demonstrates the balance between initial and novel variety recognition. From the 3-variety to the 13-variety expansion setting, the accuracy of the base varieties decreased by 10.01 percentage points, whereas that of the newly introduced varieties decreased by 11.76 percentage points, resulting in a difference of only 1.75 percentage points. This limited gap indicates that GROW maintains a reasonable balance between base-variety retention and novel-variety adaptation, without causing disproportionate degradation of the previously supported varieties. Overall, as the GrainBank was progressively expanded, GROW was able to enlarge its recognizable variety space with relatively stable performance without retraining the deployed localization model or grain encoder.

\begin{figure}
    \centering
    \includegraphics[width=\linewidth]{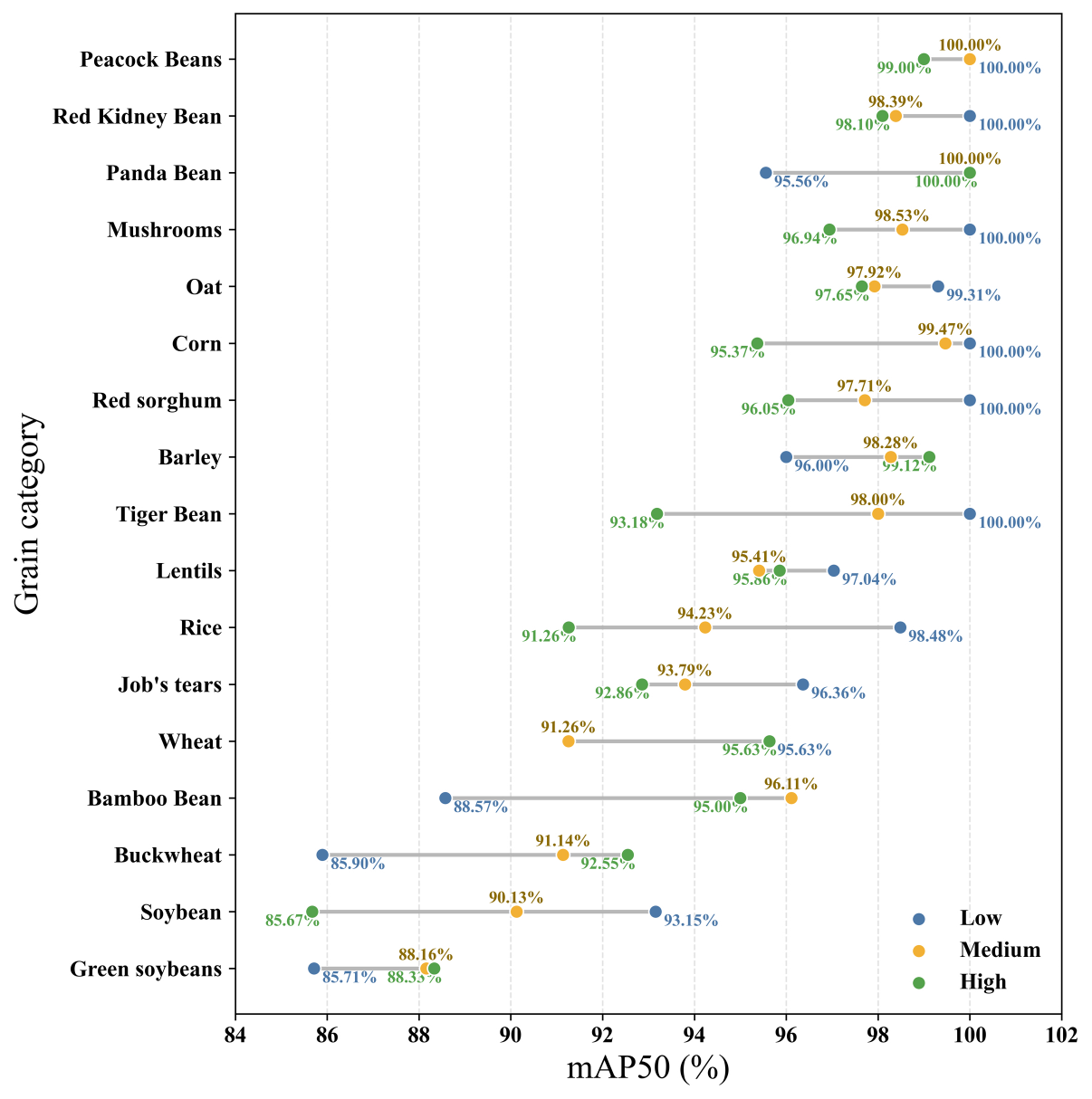}
    \caption{Single-variety mAP@50 under low, medium, and high grain-density conditions.}
    \label{fig5}
\end{figure}

\subsection{Robustness to grain-density variation}

\begin{figure}
    \centering
    \includegraphics[width=\linewidth]{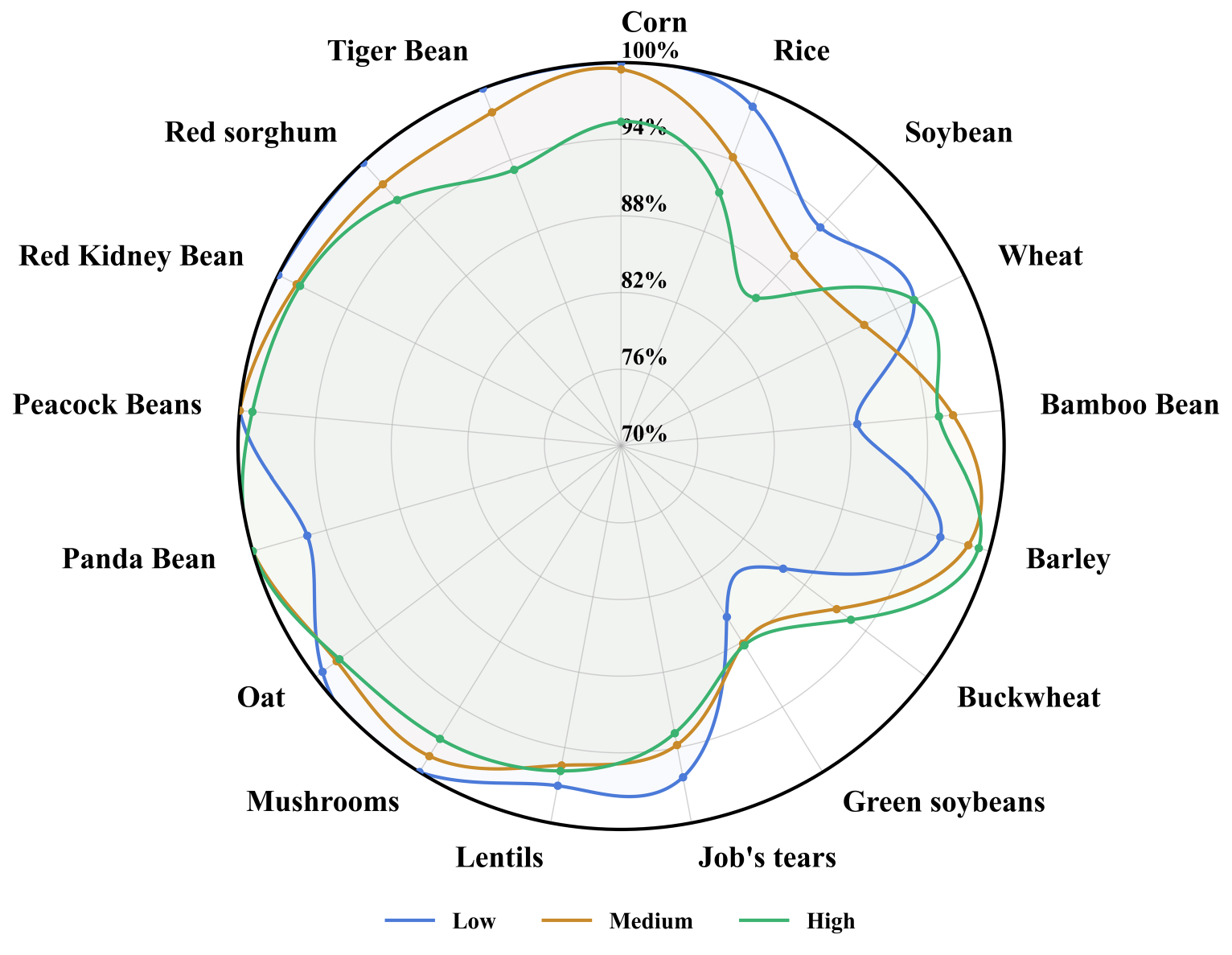}
    \caption{Radar visualization of grain-density robustness across grain varieties.}
    \label{fig6}
\end{figure}

To evaluate the recognition robustness of GROW under varying grain densities, the GrainBank category space was held fixed under a single-variety expansion setting, while only the number and spatial crowding of grain instances in the test images were varied. This experimental design was intended to examine whether density variation systematically affects class-agnostic grain localization and subsequent GrainBank retrieval. As shown in Fig.~\ref{fig5}, the sensitivity to grain density varied across varieties. Nevertheless, the similar overall
profiles of the three density levels in Fig.~\ref{fig6} indicate that GROW maintained favorable robustness to density variation. Furthermore, GROW first decomposes crowded grain images into individual grain instances and subsequently performs feature extraction and GrainBank retrieval on the cropped single-grain images. Grain density therefore primarily affects front-end instance separation and crop quality, rather than directly altering the category decision space of single-grain retrieval. Overall, class-agnostic localization reduces
the direct influence of density variation on subsequent GrainBank retrieval by decomposing multi-grain images into independent instances. The remaining local performance fluctuations may be more closely associated with grain morphology, instance adjacency, and crop quality than with a general retrieval failure caused by increased grain density.

\begin{figure*}
\centering
\includegraphics[width=\textwidth]{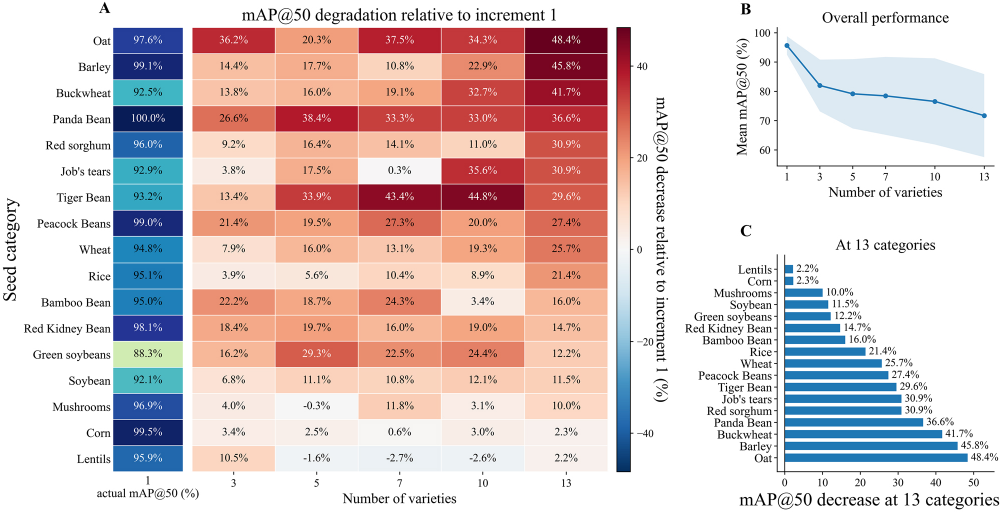}
\caption{Variety-wise performance degradation during progressive GrainBank expansion.
(A) Variety-wise mAP@50 and relative degradation across GrainBank expansion stages.
(B) Trend of mean mAP@50 with increasing GrainBank variety-set size.
(C) Ranking of variety-wise mAP@50 degradation from the initial to the final GrainBank expansion stage.}
\label{fig7}
\end{figure*}

\subsection{Variety-wise stability under progressive GrainBank expansion}

\subsubsection{Variety-wise mAP degradation}

This experiment further investigated the variety-specific effects of progressive GrainBank expansion on grain recognition performance. Unlike the overall performance evaluation, this experiment compared the expansion responses of individual varieties using single-variety test sets. Specifically, multiple GrainBanks were constructed under the 1-, 3-, 5-, 7-, 10-, and 13-variety expansion settings, with each
GrainBank containing the target variety required by the corresponding test set.

Fig.~\ref{fig7}A presents a heatmap of the mAP@50 values of individual varieties during progressive GrainBank expansion. As more varieties were introduced, the stability trajectories differed substantially across varieties. Fig.~\ref{fig7}B further summarizes the overall variety-specific performance trends across the successive expansion stages. Fig.~\ref{fig7}C ranks the varieties according to their sensitivity to retrieval competition during GrainBank expansion. Lentils and corn were the least affected by GrainBank growth, with their mAP@50 values decreasing by only 2.2 and 2.3 percentage points, respectively, after 13 varieties were introduced. However, lentils exhibited a temporary mAP@50 decrease of 10.5 percentage points at the 3-variety expansion
stage, indicating that variety-wise performance did not necessarily degrade monotonically throughout the expansion process. 

These results demonstrate marked variety-specific differences in both the magnitude and trajectory of mAP@50 degradation as the GrainBank expanded. In addition, oat and barley, which exhibited the largest performance declines, are highly similar in color and texture. This observation suggests that the effect of variety expansion depends not only on GrainBank size, but also on inter-variety representation separability and the representativeness of the stored descriptors. The resulting variety-wise ranking provides a basis for targeted
GrainBank refinement for varieties that are particularly sensitive to retrieval competition.

\subsubsection{Multi-metric evaluation of variety-wise stability}
\label{sec:quantitative_stability}

\begin{table*}[t]
\centering
\caption{Variety-wise stability metrics under retraining-free variety expansion}
\label{tab2}
\begin{tabular}{lccccl}
\hline
Variety & FRR (\%) & Degradation slope (pp/increment) & AUC over sessions (\%) & Stability category & Rank by FRR \\
\hline
Corn & 97.72 & 0.19 & 97.31 & Stable & 1 \\
Lentils & 97.70 & 0.18 & 95.32 & Stable & 2 \\
Mushrooms & 89.65 & 0.84 & 91.84 & Stable & 3 \\
Soybean & 87.47 & 0.96 & 82.40 & Stable & 4 \\
Green soybeans & 86.20 & 1.02 & 68.43 & Stable & 5 \\
Red Kidney Bean & 85.06 & 1.22 & 81.83 & Stable & 6 \\
Bamboo Bean & 83.14 & 1.33 & 80.25 & Moderate & 7 \\
Rice & 77.50 & 1.78 & 86.48 & Moderate & 8 \\
Wheat & 72.88 & 2.14 & 80.03 & Moderate & 9 \\
Peacock Beans & 72.29 & 2.29 & 78.07 & Moderate & 10 \\
Tiger Bean & 68.25 & 2.47 & 61.35 & Sensitive & 11 \\
Red sorghum & 67.81 & 2.58 & 82.22 & Sensitive & 12 \\
Job's tears & 66.72 & 2.58 & 76.48 & Sensitive & 13 \\
Panda Bean & 63.37 & 3.05 & 69.41 & Sensitive & 14 \\
Buckwheat & 54.96 & 3.47 & 70.21 & Sensitive & 15 \\
Barley & 53.82 & 3.81 & 80.06 & Sensitive & 16 \\
Oat & 50.43 & 4.03 & 65.79 & Sensitive & 17 \\
\hline
Mean & 75.00 & 2.00 & 79.26 & -- & -- \\
SD & 14.72 & 1.19 & 10.09 & -- & -- \\
\hline
\end{tabular}
\end{table*}

To further quantify variety-wise stability, three indicators were calculated from the mAP@50 degradation curves: Final Retention Rate (FRR), Class Degradation Slope, and AUC across Expansion Stages. FRR measures the retained proportion of initial performance at the maximum GrainBank expansion stage, Degradation Slope measures the average performance decline per increment, and AUC provides an integrated stability measure across all sessions.

For class $c$, these metrics are defined as:
\begin{equation}
\mathrm{FRR}_c =
\frac{\mathrm{mAP}_{c,13}}
{\mathrm{mAP}_{c,1}}
\times 100\%,
\end{equation}
\begin{equation}
\mathrm{Slope}_c =
\frac{\mathrm{mAP}_{c,1}-\mathrm{mAP}_{c,13}}
{13-1},
\end{equation}
\begin{equation}
\mathrm{AUC}_c =
\frac{1}{s_T-s_1}
\sum_{t=1}^{T-1}
\frac{\mathrm{mAP}_{c,s_t}+\mathrm{mAP}_{c,s_{t+1}}}{2}
(s_{t+1}-s_t),
\end{equation}
where $s_t \in \{1,3,5,7,10,13\}$ denotes the incremental session index.

Based on FRR, variety stability was divided into three categories: stable, $\mathrm{FRR} \geq 85\%$; moderate, $70\% \leq \mathrm{FRR} < 85\%$; and sensitive, $\mathrm{FRR} < 70\%$. As shown in Table~\ref{tab2}, the mean FRR, Degradation Slope, and AUC were 75.00\%, 2.00 percentage points per increment, and 79.26\%, respectively. These results indicate that the framework retained about three-quarters of its initial recognition performance under the largest variety expansion. However, the large standard deviation among varieties suggests that the performance loss was not uniform, but mainly concentrated in specific fine-grained categories.

Variety-wise stability was closely related to feature separability. Corn, Lentils, and Mushrooms remained stable across incremental sessions, indicating that their visual and morphological features were sufficiently distinctive in the expanded GrainBank. In contrast, Oat, Barley, and Buckwheat showed clear degradation, with FRR values close to or below 55\%. These sensitive varieties are more likely to share elongated shapes, similar colors, or weak contour differences with other grains, which increases retrieval interference during retrieval.

AUC further complements FRR by reflecting the degradation process across all incremental sessions. For example, Barley had a low final FRR but retained an AUC of 80.06\%, suggesting a gradual decline rather than an abrupt failure at early stages. By contrast, categories with low AUC were more sensitive throughout the expansion process. Overall, FRR, Degradation Slope, and AUC consistently show that incremental degradation is category dependent. Therefore, future optimization should focus on sensitive varieties by improving descriptor representativeness, descriptor discriminability, and GrainBank composition.

\subsection{Cross-background generalization and GrainBank-based recovery}

\subsubsection{Effect of background domain shift}

\begin{figure*}
    \centering
    \includegraphics[width=\textwidth]{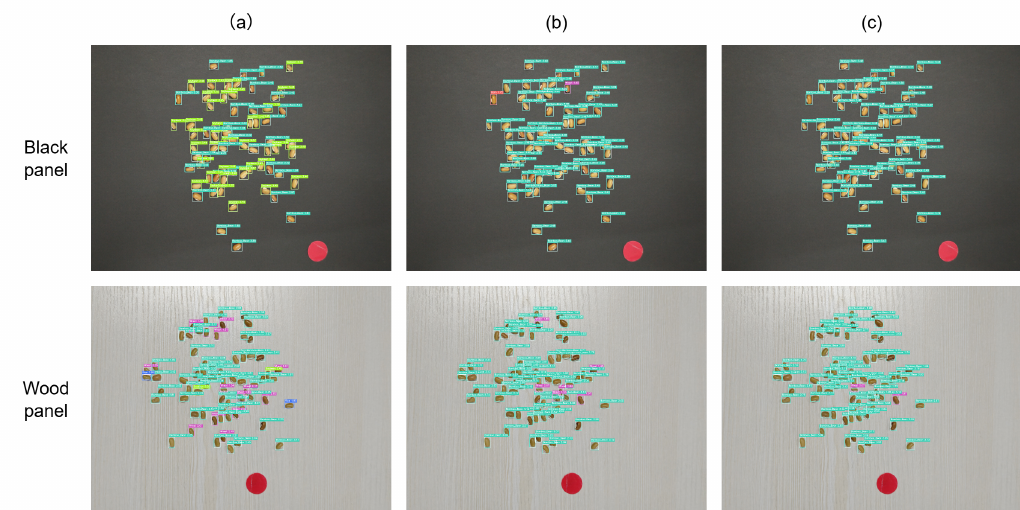}
\caption{Qualitative cross-background recognition results under different proportions of target-background descriptors in the GrainBank.
(a) GrainBank containing 0\% target-background descriptors.
(b) GrainBank containing 50\% target-background descriptors.
(c) GrainBank containing 100\% target-background descriptors.}
    \label{fig8}
\end{figure*}

\begin{figure}
    \centering
    \includegraphics[width=\linewidth]{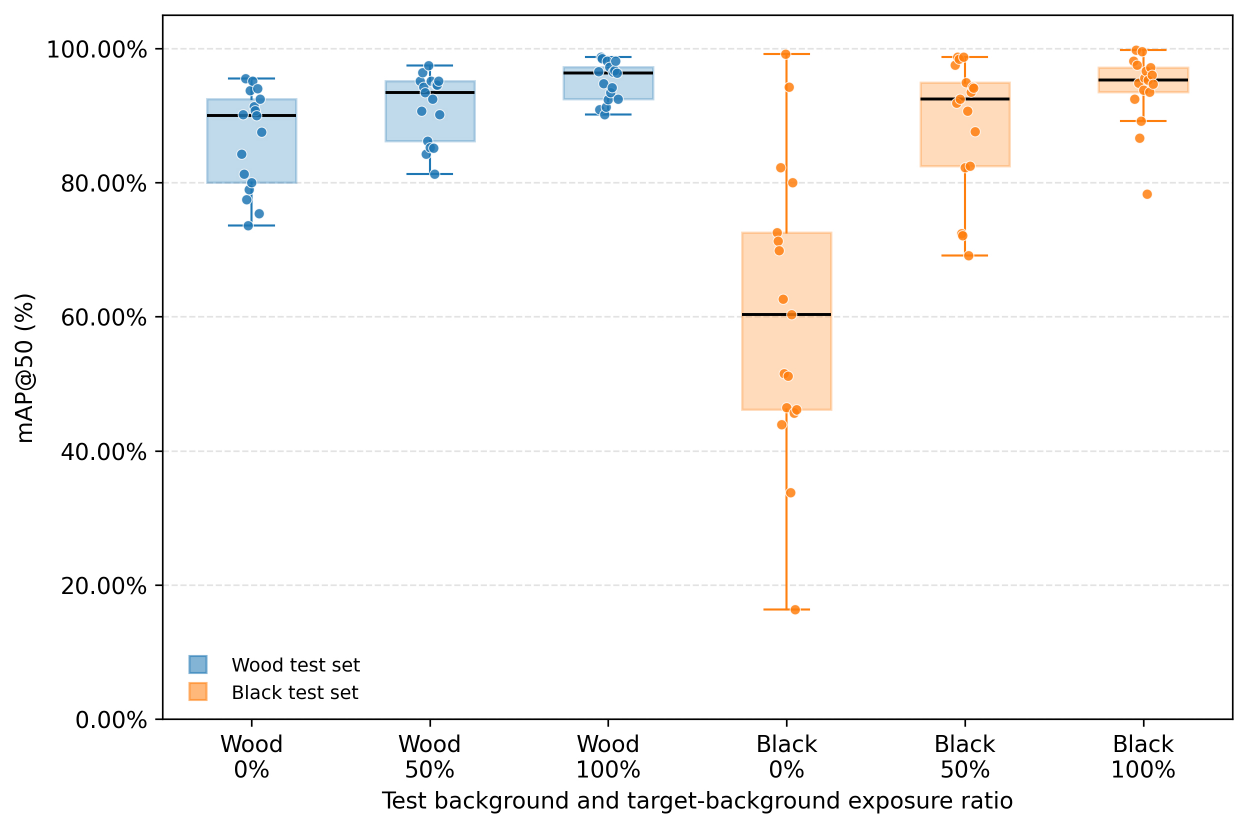}
    \caption{Cross-background recognition performance under different proportions of target-background descriptors in the GrainBank.}
    \label{fig9}
\end{figure}

The preceding experiments evaluated the effects of variety expansion and grain density on the GROW framework. This experiment further examined the cross-background generalization of GROW under background
domain shift and the feasibility of recovering recognition performance by modifying the GrainBank composition. White-background images were treated as the source domain, whereas black and wood-textured images were used as the target domains. The proportion of target-background descriptors stored in the GrainBank was set to 0\%, 50\%, and 100\%, while the class-agnostic localization model and grain encoder remained fixed. This design allowed the observed performance changes to be attributed primarily to background consistency between the query instances and the GrainBank composition.

As shown in fig~\ref{fig8}, the qualitative examples illustrate the recognition results for Bamboo Bean under the two target-background conditions as the proportion of matched-background descriptors in the GrainBank increased from 0\% to 100\%. When the target-background proportion reached 100\%, all grain instances in the black-background test image were correctly recognized, whereas only two instances in the wood-textured image were misclassified as wheat. The box-and-scatter plots in fig~\ref{fig9} further show that the black-background test set was more sensitive to changes in GrainBank background composition. Nevertheless, recognition performance under both target backgrounds recovered substantially as the proportion of target-background descriptors increased. 

\subsubsection{Quantitative assessment of background-induced degradation and GrainBank-based recovery}
\label{sec:background_recovery}
To quantitatively characterize background-induced performance degradation and retraining-free recovery through GrainBank updates, three complementary metrics were calculated: Background Performance
Drop (BPD), GrainBank Adaptation Gain (GAG), and Domain Recovery Ratio (DRR). Let $M_{c}^{WW}$ denote the mAP@50 of variety $c$ when both the GrainBank and test set use a white background. Let $M_{c}^{WT}$ denote
the performance obtained when a white-background GrainBank is directly evaluated on a target-background test set. Furthermore, $M_{c}^{50T}$ and $M_{c}^{100T}$ denote the performance when target-background descriptors account for 50\% and 100\% of the GrainBank, respectively, with evaluation conducted on the corresponding target background. The metrics are defined as
\begin{equation}
\mathrm{BPD}_c
=
M_{c}^{WW}-M_{c}^{WT},
\end{equation}
\begin{equation}
\mathrm{GAG50}_c
=
M_{c}^{50T}-M_{c}^{WT},
\end{equation}
\begin{equation}
\mathrm{GAG100}_c
=
M_{c}^{100T}-M_{c}^{WT},
\end{equation}
\begin{equation}
\mathrm{DRR50}_c
=
\frac{\mathrm{GAG50}_c}{\mathrm{BPD}_c}
\times100\%,
\quad \mathrm{BPD}_c>0,
\end{equation}
and
\begin{equation}
\mathrm{DRR100}_c
=
\frac{\mathrm{GAG100}_c}{\mathrm{BPD}_c}
\times100\%,
\quad \mathrm{BPD}_c>0.
\end{equation}

BPD quantifies the absolute mAP@50 loss caused by the mismatch between the white-background GrainBank and the target-background queries. GAG measures the absolute performance recovered after target-background
descriptors are incorporated into the GrainBank. DRR further normalizes this gain by the original background-induced loss and therefore represents the proportion of lost performance recovered through GrainBank adaptation. Because DRR is meaningful only when $\mathrm{BPD}_c>0$, the corresponding number of eligible varieties is also reported in Table~\ref{tab3}.

\begin{figure}
    \centering
    \includegraphics[width=0.8\linewidth]{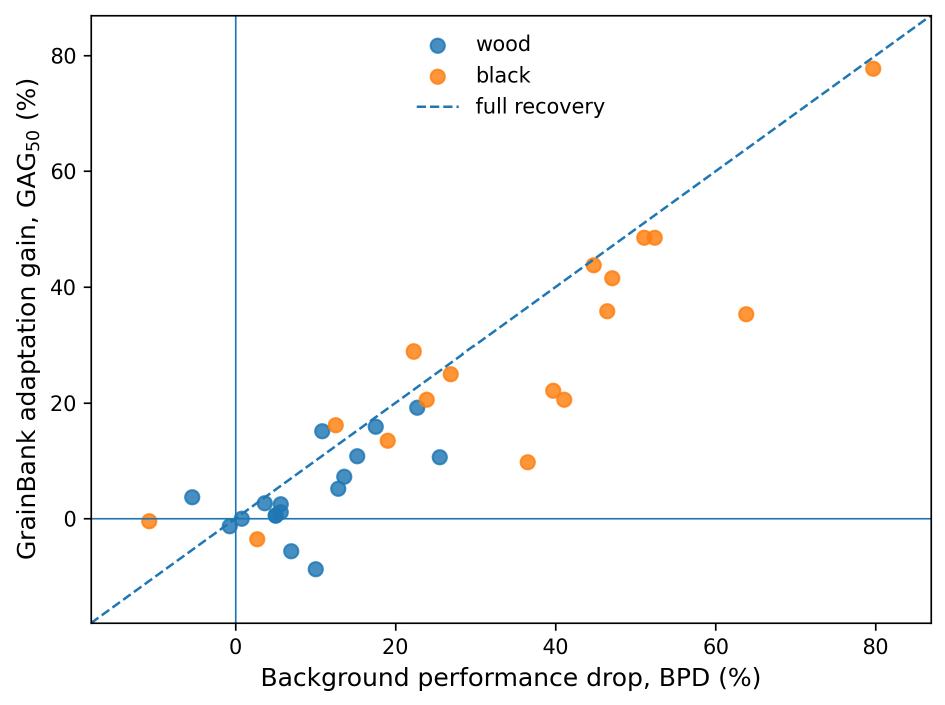}
    \caption{Relationship between Background Performance Drop (BPD) and GrainBank Adaptation Gain obtained with a GrainBank containing 50\% target-background descriptors (GAG$_{50}$).}
    \label{fig10}
\end{figure}

Fig~\ref{fig10} further illustrates the relationship between BPD and GAG50. Samples from the black-background condition are generally located further to the right and higher on the vertical axis, indicating both stronger background-induced degradation and larger adaptation gains after incorporating target-background descriptors. Points close to the dashed identity line indicate that the gain obtained with a 50\% target-background GrainBank approximately compensated for the original performance loss. Points below this line represent incomplete recovery, which may be associated with localization and crop quality, inter-variety descriptor similarity, or insufficient representativeness of the stored GrainBank descriptors.

Overall, although background domain shift caused clear cross-background performance degradation, most of the lost recognition performance could be recovered by rapidly updating the GrainBank while keeping the class-agnostic localization model and grain encoder fixed.

\begin{table*}[t]
\centering
\caption{Quantitative summary of background-induced degradation and
GrainBank-based recovery under different target backgrounds.}
\label{tab3}
\resizebox{\textwidth}{!}{
\begin{tabular}{lcccccccccc}
\hline
Target background &
Varieties with BPD$>0$ &
$M^{WW}$ (\%) &
$M^{WT}$ (\%) &
$M^{50T}$ (\%) &
$M^{100T}$ (\%) &
BPD (pp) &
GAG$_{50}$ (pp) &
GAG$_{100}$ (pp) &
DRR$_{50}$ (\%) &
DRR$_{100}$ (\%) \\
\hline
Wood-textured &
15 &
$95.65 \pm 3.16$ &
$86.56 \pm 7.38$ &
$91.26 \pm 4.99$ &
$95.08 \pm 2.87$ &
$9.09 \pm 8.18$ &
$4.70 \pm 7.61$ &
$8.52 \pm 6.87$ &
$34.63 \pm 60.30$ &
$93.01 \pm 63.19$ \\

Black &
16 &
$95.65 \pm 3.16$ &
$60.43 \pm 21.74$ &
$88.91 \pm 9.78$ &
$94.06 \pm 5.29$ &
$35.22 \pm 22.46$ &
$28.48 \pm 20.16$ &
$33.63 \pm 20.23$ &
$69.74 \pm 59.95$ &
$94.93 \pm 17.62$ \\

Combined &
31 &
$95.65 \pm 3.11$ &
$73.50 \pm 20.77$ &
$90.09 \pm 7.74$ &
$94.57 \pm 4.22$ &
$22.16 \pm 21.28$ &
$16.59 \pm 19.26$ &
$21.08 \pm 19.59$ &
$52.75 \pm 61.74$ &
$94.00 \pm 44.94$ \\
\hline
\end{tabular}}
\end{table*}

\begin{table*}[t]
\centering
\caption{Component-level detection performance on initial-variety and novel-variety grain test sets.}
\label{tab4}
\resizebox{\textwidth}{!}{
\begin{tabular}{llccccccccc}
\hline
Evaluation set & Images & Precision (\%) & Recall (\%) & F1-score (\%) & mAP@50 (\%) & mAP@75 (\%) & mAP@50:95 (\%) & Count MAE / RMSE \\
\hline
Initial-variety grains & 273 & 97.79 & 96.75 & 97.27 & 96.46 & 88.87 & 75.21 & 2.49 / 8.22 \\
Novel-variety grains & 245 & 98.68 & 98.71 & 98.69 & 97.65 & 80.89 & 63.83 & 0.23 / 0.65 \\
\hline
\end{tabular}}
\end{table*}

\begin{table*}[!b]
\centering
\caption{Variety-wise classification performance of the ConvNeXt encoder on initial-variety grains.}
\label{tab5}
\resizebox{\textwidth}{!}{
\begin{tabular}{lcccccc}
\hline
Variety & GT support & Pred. support & Precision (\%) & Recall (\%) & F1-score (\%) & AP/mAP@50 (\%) \\
\hline
Corn & 573 & 572 & 100.00 & 99.83 & 99.91 & 99.83 \\
Rice & 675 & 688 & 98.11 & 100.00 & 99.05 & 98.24 \\
Soybean & 415 & 418 & 99.28 & 100.00 & 99.64 & 99.80 \\
Wheat & 1115 & 1104 & 99.91 & 98.92 & 99.41 & 98.92 \\
\hline
Overall & 2778 & 2782 & 99.33 & 99.69 & Micro-F1: 99.46 / Macro-F1: 99.50 & 99.19 \\
\hline
\end{tabular}}
\end{table*}

\subsection{Ablation studies}

\subsubsection{Component-level validation of localization and visual encoding}

\begin{figure}
\centering
\includegraphics[width=\linewidth]{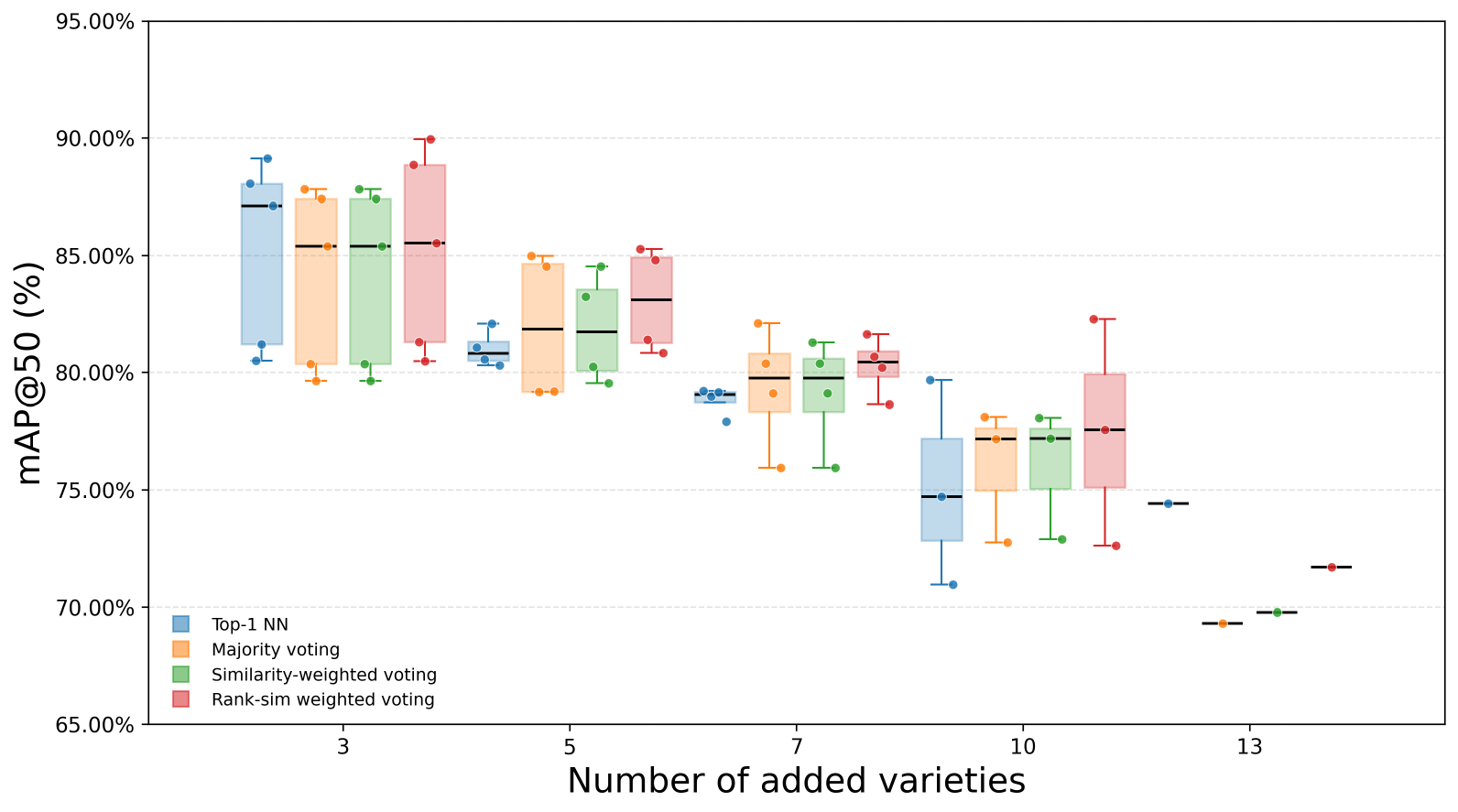}
\caption{Comparison of mAP@50 among different retrieval voting strategies during progressive GrainBank expansion.}
\label{fig11}
\end{figure}

\begin{table}[htbp]
\centering
\caption{Definitions of the compared retrieval voting strategies}
\label{tab7}
\small
\setlength{\tabcolsep}{3pt}
\renewcommand{\arraystretch}{1.08}

\begin{tabularx}{\columnwidth}{@{}
>{\raggedright\arraybackslash}X
>{\centering\arraybackslash}p{0.10\columnwidth}
>{\centering\arraybackslash}p{0.37\columnwidth}
@{}}
\hline
Retrieval strategy & Top-$k$ & Vote definition \\
\hline
Top-$1$ NN
& $1$
& nearest stored descriptor \\

Majority top-$k$
& $5$
& $v=1$ \\

Similarity-weighted top-$k$
& $5$
& $v=[\max(s,0)]^{\gamma}$ \\

Rank-similarity weighted top-$k$
& $5$
& $v=\lambda^{r}[\max(s,0)]^{\gamma}$ \\
\hline
\end{tabularx}
\end{table}

\begin{figure*}
\centering
\includegraphics[width=\textwidth]{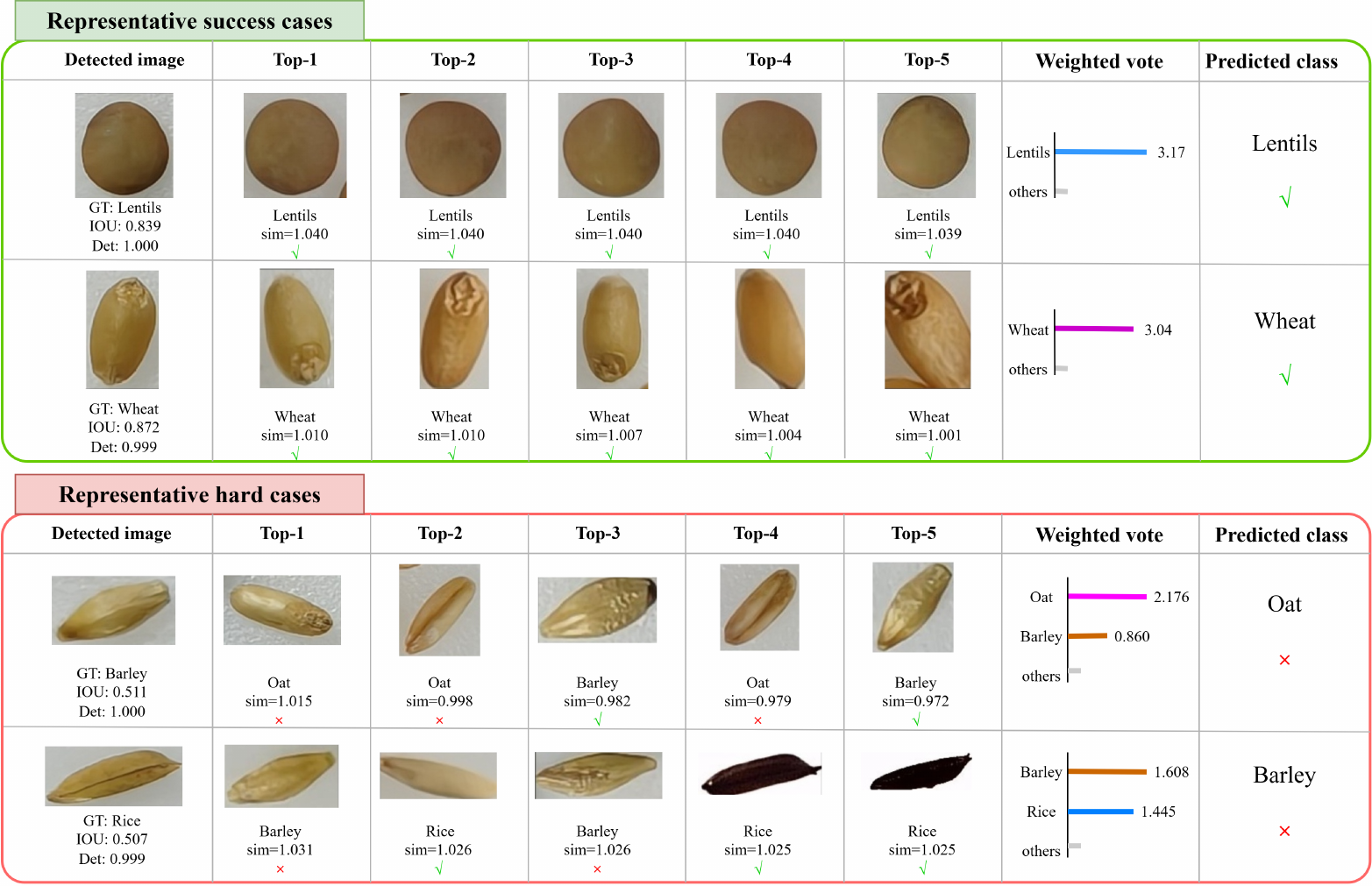}
\caption{Representative successful and failure cases of rank- and similarity-weighted top-$k$ voting for GrainBank-based variety recognition.}
\label{fig12}
\end{figure*}

To validate the reliability of the core components of GROW, this experiment separately evaluated the localization performance of the class-agnostic detector on base and newly introduced varieties, as well
as the closed-set variety discrimination capability of the ConvNeXt encoder. Table~\ref{tab4} compares the detection performance on the two variety groups. Notably, the detector was trained exclusively using images from the base varieties. The newly introduced varieties achieved an mAP@50 that was 1.19 percentage points higher than that of the base varieties. By contrast, the base varieties outperformed the newly introduced
varieties by 11.38 percentage points in mAP@50:95. These results indicate that grain varieties not involved in detector training can still be reliably localized and extracted as individual instances by the class-agnostic localization model. However, under the stricter mAP@50:95 criterion, the lower performance on newly introduced varieties suggests reduced bounding-box regression accuracy for previously unseen grain appearances.

The closed-set evaluation results of the encoding module are reported in Table~\ref{tab5}. Under the closed-set setting, the visual encoder learned discriminative representations among the known varieties, confirming
that the extracted visual embeddings provide a reliable feature basis for subsequent GrainBank retrieval.

Overall, the class-agnostic localization model maintained stable localization performance on both base and newly introduced varieties, while ConvNeXt exhibited strong variety discrimination under closed-set
evaluation. These results validate the modular design of GROW at the component level: the front-end localization module extracts individual grain instances without relying on variety labels, whereas the visual
encoder provides reliable representations for subsequent descriptor fusion and GrainBank retrieval, thereby supporting retraining-free variety expansion.

\subsubsection{Effect of weighted top-$k$ retrieval}

To evaluate the contribution of the retrieval decision mechanism to the recognition performance of GROW, different top-$k$ voting strategies were compared while keeping the class-agnostic localization
model, grain encoder, and GrainBank composition unchanged. As shown in Fig~\ref{fig11}, recognition accuracy was evaluated under four voting mechanisms, whose definitions are summarized in Table~\ref{tab7}. The Top-1
nearest-neighbor (Top-1NN) rule assigns the label of the highest-scoring retrieved instance as the final prediction. However, this strategy directly produces an incorrect prediction when the top-ranked neighbor
belongs to an incorrect variety. Majority voting was subsequently introduced to aggregate multiple retrieved neighbors, but the results showed that it could be dominated by a larger number of low-similarity
mismatches. Based on these observations, the rank--similarity weighted top-$k$ voting strategy was adopted as the final retrieval decision mechanism.

As illustrated in Fig~\ref{fig12}, the final prediction under the proposed strategy is jointly determined by the top five retrieved neighbors. Specifically, retrieval rank and descriptor similarity are jointly incorporated to reduce the influence of low-confidence matches while strengthening the variety evidence provided by highly ranked and strongly matched neighbors. These results validate the effectiveness of rank--similarity weighted voting for GrainBank-based retrieval and provide a more reliable decision layer for maintaining stable recognition as the GrainBank category space expands.

\subsubsection{Effect of morphological physical descriptors}

\begin{figure}
    \centering
    \includegraphics[width=\linewidth]{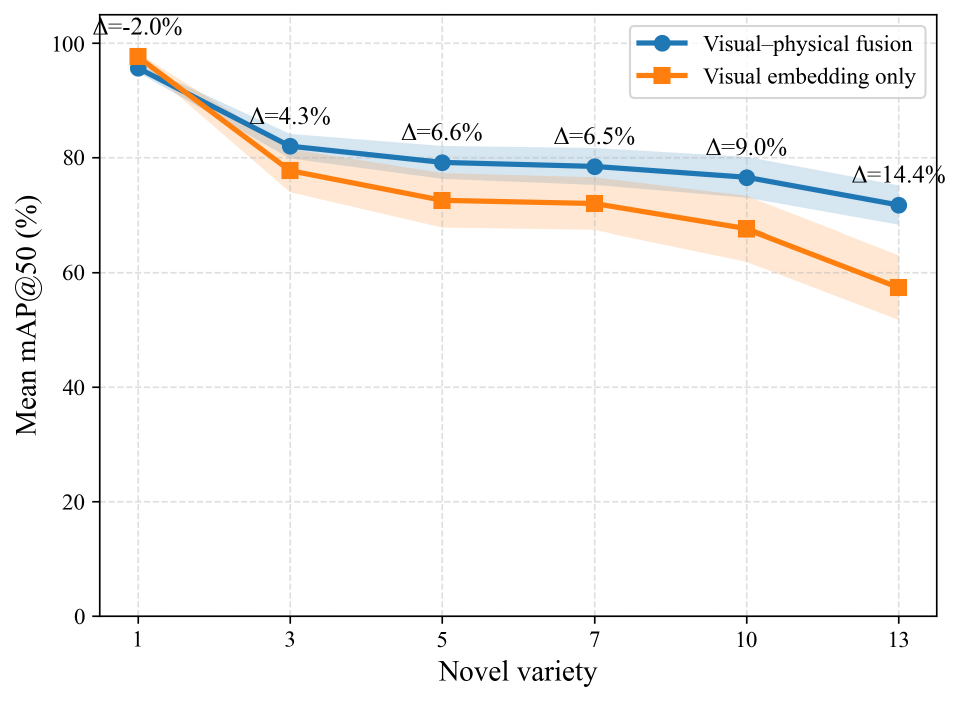}
    \caption{Overall effect of visual--morphological descriptor fusion across GrainBank expansion stages.}
    \label{fig13}
\end{figure}

The primary recognition cue in GROW remains the visual embedding extracted by ConvNeXt. Morphological priors are not intended to replace visual features; instead, they introduce complementary shape constraints through descriptors such as aspect ratio, extent, circularity, solidity, relative area, and relative perimeter. As shown in Fig~\ref{fig13}, recognition performance improved after incorporating the morphological priors while the localization model, visual encoder, GrainBank composition, and voting mechanism were kept unchanged. This
controlled comparison indicates that the observed gains arose from visual--morphological descriptor fusion rather than changes in other components.

\begin{figure}
    \centering
    \includegraphics[width=\linewidth]{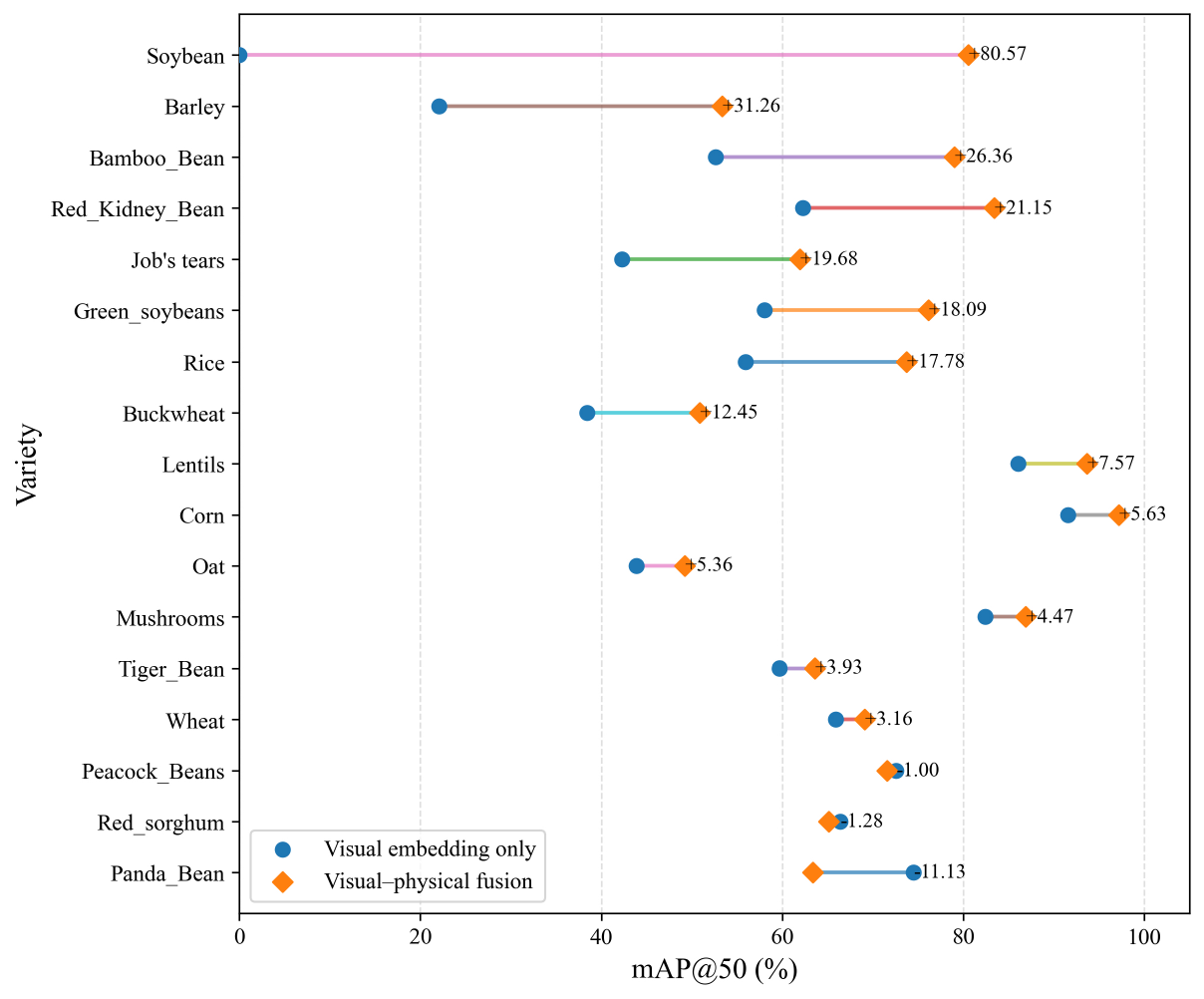}
    \caption{Variety-wise comparison between visual embeddings alone and visual--morphological descriptor fusion at the final GrainBank expansion stage.}
    \label{fig14}
\end{figure}

Fig~\ref{fig14} further presents the variety-specific effects of the morphological priors, revealing a pronounced difference in their contributions across varieties. Several varieties, including soybean and barley, achieved substantial improvements, broadly consistent with the intended role of the morphological descriptors. Under a visual-only retrieval setting, variety pairs such as soybean and lentils, as well as barley, rice, and oat, exhibit highly similar color and texture characteristics. Nevertheless, they retain subtle differences in morphological properties, including aspect ratio and circularity. Integrating these morphological priors with the visual embeddings therefore provides complementary constraints for distinguishing visually similar varieties. These results validate the effectiveness of visual--morphological descriptor fusion in GrainBank-based retrieval and provide additional support for maintaining stable recognition as the GROW category space expands.

\subsubsection{Comparison with joint retraining under progressive variety expansion}

To further validate the effectiveness of GROW for category expansion, the proposed framework was compared with conventional joint retraining (JR). At each expansion stage, both methods used the same combination
of newly introduced varieties and the same 150 images per new variety, while all other relevant experimental settings were kept consistent. The key difference was that GROW only extracted descriptors from the
new-variety samples and rapidly appended them to the GrainBank, whereas JR merged the base- and new-variety samples and retrained the ConvNeXt classification model.

\begin{figure}
    \centering
    \includegraphics[width=\linewidth]{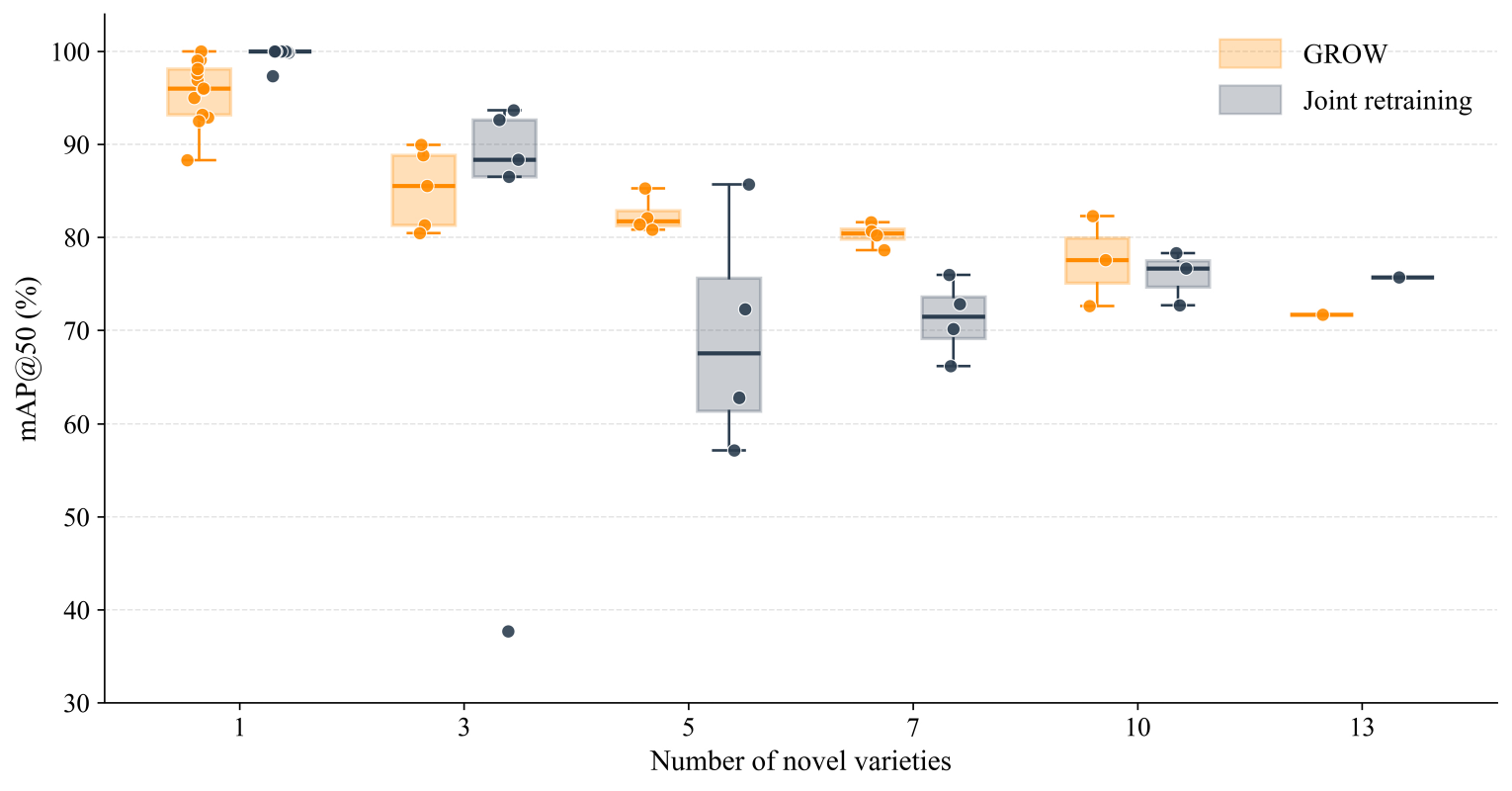}
    \caption{Comparison of mAP@50 between GROW and joint retraining under progressive variety expansion.}
    \label{fig15}
\end{figure}

Fig~\ref{fig15} compares the mAP@50 values of the two methods when 1, 3, 5, 7, 10, and 13 varieties were newly introduced. With only one new variety, JR achieved a mean mAP@50 of 99.78\%, clearly exceeding the
95.73\% obtained by GROW. However, as the number of newly introduced varieties increased, JR exhibited pronounced performance degradation and substantially greater variation, with several category combinations
resulting in severe performance collapse. In the 3-variety expansion experiment, one JR trial achieved an mAP@50 of only 37.69\%. The newly introduced varieties in this trial were Bamboo Bean, lentils, and oat,
which share similar color and texture characteristics with several base varieties and likely contributed substantially to the observed degradation. Under the same category combination, GROW achieved an
mAP@50 of 81.30\%, substantially outperforming JR. Overall, GROW exhibited a more gradual reduction in mAP@50 as the recognizable variety space expanded, whereas JR showed both sharper degradation and
larger cross-combination fluctuations. These results indicate that joint retraining is more sensitive to category composition and sample distribution, while GROW provides more stable adaptation across different variety combinations during medium-scale category expansion.

The update efficiency of the two methods was also evaluated. All experiments were conducted on the same computer, with JR trained for 30 epochs using a batch size of 64 and a learning rate of 0.0005. At
each expansion stage, JR required the ConvNeXt classification model to be retrained, exported, and redeployed, resulting in an average update time of 4153~s. By contrast, GROW only required feature extraction for the newly introduced varieties and descriptor appending to the GrainBank, with an average update time of 39~s. Thus, without accessing the base-variety training data, updating model parameters, or
redeploying the recognition model, GROW achieved recognition performance comparable to or higher than that of joint retraining while reducing the category-registration time by approximately two orders of magnitude.

\section{Discussion}

\subsection{Framework flexibility and deployment relevance}

Reliable instance localization and GrainBank updating enabled the proposed framework to expand the recognizable variety space without retraining the full model. This indicates that grain recognition does not necessarily need to rely on a fixed classifier head. Instead, GrainBank-based retrieval provides a more flexible mechanism when the category space changes over time. Compared with previous hyperspectral open-world maize seed recognition studies \citep{zhangVisNIRHyperspectralImaging2022,zhangOpenSetMaize2023a}, the present work uses RGB mixed-grain images and performs instance-level localization, recognition, and counting, which is more suitable for low-cost machine-vision deployment.

\subsection{Scalability under progressive GrainBank expansion}

Under increasing mixed-grain category complexity, GROW demonstrated practical scalability through progressive GrainBank expansion. When a small number of novel varieties were introduced, the framework maintained balanced performance between the base and newly introduced varieties. With further expansion, the overall recognition performance gradually declined because the increasing candidate variety space intensified inter-variety competition and reduced descriptor separability \citep{rebuffiICaRL2017,sun2026lure}. Therefore, the performance degradation observed under larger GrainBank scales should be interpreted primarily as retrieval interference among similar descriptors rather than as parameter forgetting or a failure of the framework itself.

\subsubsection{Role of morphological priors in GrainBank retrieval}

Visual-morphological fusion demonstrated that morphological priors can serve as useful complementary constraints for GrainBank-based recognition. Although morphological priors slightly reduced performance
at the initial GrainBank expansion stage, their advantage became more evident as the number of supported varieties increased. This is reasonable because visual embeddings may provide sufficient discriminability when the variety space is relatively simple, whereas shape descriptors become more valuable when visually similar varieties are introduced. This finding agrees with previous seed morphometry studies, which showed that geometric traits such as size, contour, and shape descriptors can support seed classification and phenotyping \citep{dubeyPotentialArtificialNeural2006,dayrellAutomatedExtractionSeed2023}. However, because these descriptors rely on threshold-based contour extraction, grain adhesion, shadows, or incomplete crops may introduce measurement noise. Therefore, morphological priors should be treated as complementary constraints rather than dominant recognition features.

\subsubsection{Robustness to grain density and background shift}

Density variation did not lead to a global monotonic performance decline, suggesting that the detector and retrieval modules were relatively robust to different grain-density conditions. Local fluctuations still occurred for categories with similar morphology or boundary adhesion, which is consistent with previous detection-based grain counting studies \citep{maWheatSeedDetection2024a,zouRiceGrainDetection2023}. In contrast, background shift had a stronger effect than density variation, consistent with previously reported
cross-environment degradation in agricultural vision\citep{leonBroadleafDomain2024}. In particular, the black-background test set caused a much larger performance drop than the wood-background test set, indicating that background appearance can still affect cropped-grain representations. Nevertheless, adding a target-background stored descriptor substantially recovered the lost performance, suggesting that stored descriptor updating can also serve as a lightweight domain adaptation strategy.

\subsection{Engineering implications of GROW}

GROW demonstrates that, in agricultural vision tasks where object appearance and structure remain relatively stable while the supported category set continues to evolve, newly introduced categories do not necessarily require model retraining. By storing variable variety-specific information in the GrainBank, GROW enables novel varieties to be registered through representative samples and descriptor appending, thereby reducing the burdens of repeated mixed-scene annotation, model retraining, and terminal redeployment. In this sense, GrainBank also provides a lightweight digital reference for organizing reusable variety descriptors and morphological traits, consistent with the development of digital seed phenotyping \citep{songSeedImagingOmics2026}. However, several limitations remain. First, the current evaluation covers a moderate variety scale, and further validation is required as the GrainBank expands to substantially larger category spaces. Second, newly introduced varieties still require representative labeled samples for GrainBank registration, and completely unregistered categories are not explicitly rejected. In addition, exhaustive descriptor matching may increase retrieval cost as the GrainBank grows. Future work will therefore focus on large-scale GrainBank organization, efficient retrieval, and explicit unknown-category handling.

\section{Conclusion}
In this work, we propose GROW, a framework for scalable grain recognition and quantitative analysis. 
GROW first performs class-agnostic grain localization, which reformulates multi-category,
multi-grain detection as single-class localization of individual grains, reducing category-related interference and neighboring-grain interference for greater robustness while remaining applicable to novel varieties. 
For each localized grain, GROW integrates visual information with grain morphological traits to construct a unified grain feature, which is organized in a GrainBank for inference. Instead of predicting varieties through a predefined classifier, GROW determines each query
grain by matching its feature against those stored in the GrainBank.
Novel varieties can therefore be incorporated by adding their grain features from a small set of grain images, without modifying the deployed model. 
 Compared with joint retraining, GROW reduced the average category-registration time from 4153~s to only 39~s, demonstrating its substantial advantage in low-cost and efficient variety expansion.

\section*{Code availability}
https://github.com/Hehenum17/GROW

\section*{Funding}

This work was supported by grants from the National Natural Science Foundation of China (32270431), the National Science and Technology Major Project of China (2022ZD0115705), Key Research and Development Plan of Hubei Province (2022BBA0045), the Fundamental Research Funds for the Central Universities (2662024GXPY002).

\section*{Declaration of competing interest}

The authors declare that they have no known competing financial
interests or personal relationships that could have appeared to
influence the work reported in this paper.

\printcredits

\section*{Data availability}
Data will be made available on request.

\bibliographystyle{cas-model2-names}

\bibliography{cas-refs}

\end{document}